\documentclass{article}
\usepackage{ais25}
\usepackage[T1]{fontenc}

\usepackage[table, dvipsnames]{xcolor}
\usepackage{times}
\usepackage{graphicx}
\usepackage{booktabs} 
\usepackage{multirow}
\usepackage{amsmath}
\usepackage{mathtools}
\usepackage{subcaption}
\usepackage{amssymb}
\usepackage{url}
\usepackage{hyperref}
\hypersetup{
    citecolor=CadetBlue,
    colorlinks=true,
    linkcolor=CadetBlue,
    filecolor=magenta,      
    urlcolor=cyan}

\usepackage{xspace}
\newcommand{\method}[1]{\texttt{#1}\xspace}
\newcommand{\FIM}{\method{FIM-PP}}

\newcommand{\nomath}[1]{}

\newcommand{\markel}{\kappa}
\newcommand{\history}{\mathcal{H}}
\newcommand{\kernel}{\gamma}
\newcommand{\baseint}{\mu}
\newcommand{\contextseq}{\mathcal{S}}
\newcommand{\contextset}{\mathcal{C}}
\newcommand{\hawfactorel}{z}

\newcommand{\hist}{\text{\tiny{hist}}}

\newcommand{\histencfinal}{\mathbf{h}^\hist}
\newcommand{\muhist}{\hat{\mu}_{\markel^\prime}^{\history_t}}
\newcommand{\alphahist}{\hat{\alpha}_{\markel^\prime}^{\history_t}}
\newcommand{\betahist}{\hat{\beta}_{\markel^\prime}^{\history_t}}

\title{On Foundation Models for Temporal Point Processes to Accelerate Scientific Discovery}
\author{
\name{David Berghaus\textsuperscript{1,2}, Patrick Seifner\textsuperscript{1,3}, \name{Kostadin Cvejoski}\textsuperscript{1,4}, \\ \& Ramsés J. Sánchez\textsuperscript{1,2,3}} \\ \addr{\textsuperscript{1}Lamarr Institute, \textsuperscript{2}Fraunhofer IAIS, \textsuperscript{3}University of Bonn, \textsuperscript{4}JetBrains Research} 
}

\begin{document}
\maketitle

\begin{abstract}
Many scientific fields, from medicine to seismology, rely on analyzing sequences of events over time to understand complex systems. Traditionally, machine learning models must be built and trained from scratch for each new dataset, which is a slow and costly process.
We introduce a new approach: a single, powerful model that learns the underlying patterns of event data \textit{in-context}. We trained this "foundation model" on millions of simulated event sequences, teaching it a general-purpose understanding of how events can unfold.
As a result, our model can analyze new scientific data instantly, without retraining, simply by looking at a few examples from the dataset. It can also be quickly fine-tuned for even higher accuracy. This approach makes sophisticated event analysis more accessible and accelerates the pace of scientific discovery. 
Our pretrained model, repository and tutorials \textbf{will soon be available online}\footnote{\url{https://fim4science.github.io/OpenFIM/intro.html}}.
\end{abstract}


\section{Introduction and Related Work}
\label{sec:intro}
The pace of scientific discovery is being dramatically accelerated by artificial intelligence, with breakthrough applications emerging across numerous fields. A landmark example is AlphaFold \citep{jumper2021highly}, which has revolutionized structural biology by accurately predicting protein structures. 

This success is part of a broader trend where AI is becoming an indispensable tool for researchers, enabling them to tackle previously intractable problems in fields like weather-forecasting \citep{Price2025Probabilistic}, material science \citep{Merchant2023}, medicine \citep{Poplin2018,enguehard2020neuraltemporalpointprocesses,berger2,berger,berger3} and mathematics \citep{novikov2025alphaevolvecodingagentscientific,ramsey}. 

A common challenge in these and many other scientific domains is the need to model \textit{sequences of discrete events that occur at irregular intervals in continuous time} \citep{cvejoski2020recurrent, dynamicreview2021,ojeda2021learning,pmlr-v202-seifner23a,khanna2025unifiedmultimodalfinancialforecasting}. Temporal point processes (TPPs) provide a powerful and principled mathematical framework for this task \citep{daley2003introduction}. By capturing the underlying dynamics of event streams --- not just \textit{when} events happen, but also \textit{what} they are --- TPPs enable us to forecast future events, understand complex dependencies, and gain deeper insights into the phenomena being studied. 

For this reason, various works have modeled TPPs using deep learning techniques \citep{du2016recurrent,mei-2017-neural-hawkes,kim2019attentive,panos-2023-vi-dpp}.

However, a fundamental limitation of these approaches is their \textit{lack of transferability}: a new TPP model must be trained from scratch for each new dataset, a slow and costly process. To overcome this, a new paradigm has emerged: pretraining a single, powerful model on vast amounts of synthetic data that can then perform inference \textit{in-context} (or \textit{zero-shot}) on new, unseen problems. Within this paradigm, two variants have recently emerged: Prior-fitted Networks (PFNs) and Foundation Inference Models (FIMs). PFNs train networks to approximate \textit{predictive posterior distributions} in a sequence-to-sequence fashion, often implemented with transformer architectures~\citep{muller2022transformers,hollmann2022tabpfn, muller2025position}. FIMs, by contrast, focus on directly \textit{estimating the infinitesimal operators} of stochastic processes, thereby retaining a degree of interpretability~\citep{fim_mjp, fim_sde, fim_imputation, ais_ode, ais_eq}.

In this work, we demonstrate the first Foundation Inference Model for Marked Temporal Point Processes, the complete details of which will be presented elsewhere~\citep{fim_pp}. 
Our model learns to directly estimate the conditional intensity function --- the fundamental operator governing TPP dynamics --- from a small context of observed event sequences. This allows practitioners to obtain accurate, \textit{zero-shot} characterizations of event dynamics, accelerating the pace of scientific discovery.

\section{In-Context TPP Inference}
\label{sec:FIM}
We propose a novel in-context learning method for MTPP intensity inference. First, we generate a large set of marked event sequences from a broad distribution over MTPPs. This yields training data for a neural network recognition model, which is trained to estimate the underlying ground-truth intensity function via maximum likelihood. Such a pretrained inference model can then be applied directly to real-world problems (zero-shot) or rapidly finetuned to target data.

\subsection{Synthetic Dataset Generation}
To design a synthetic dataset, we define a \textit{prior distribution over Hawkes processes}. An instance is characterized by a conditional intensity function for each mark $\markel$:
\begin{equation}
    \lambda (t, \markel \mid \history_t) =  \max \left(0, \baseint_\markel(t) + \sum_{(t^\prime, \markel^\prime) \in \history_t} \hawfactorel_{\markel\markel^\prime}\kernel_{\markel\markel^\prime} (t - t^\prime) \right). 
\end{equation}
We generate a diverse dataset by sampling the functional forms and parameters for the base intensities $\baseint_\markel$ and interaction kernels $\kernel_{\markel\markel^\prime}$ from broad distributions (e.g., constant, sinusoidal, exponential decay). The factors $\hawfactorel_{\markel\markel^\prime} \in \{-1, 0, 1\}$ are sampled to create excitatory, inhibitory, or neutral interactions with varying levels of sparsity. From each sampled intensity function, we simulate a collection of event sequences using Ogata's thinning algorithm \citep{ogata-1981-simulation-for-pp}.

\subsection{Foundation Inference Model Architecture}
\begin{figure}[t]
\begin{center}
\includegraphics[width=0.85\textwidth]{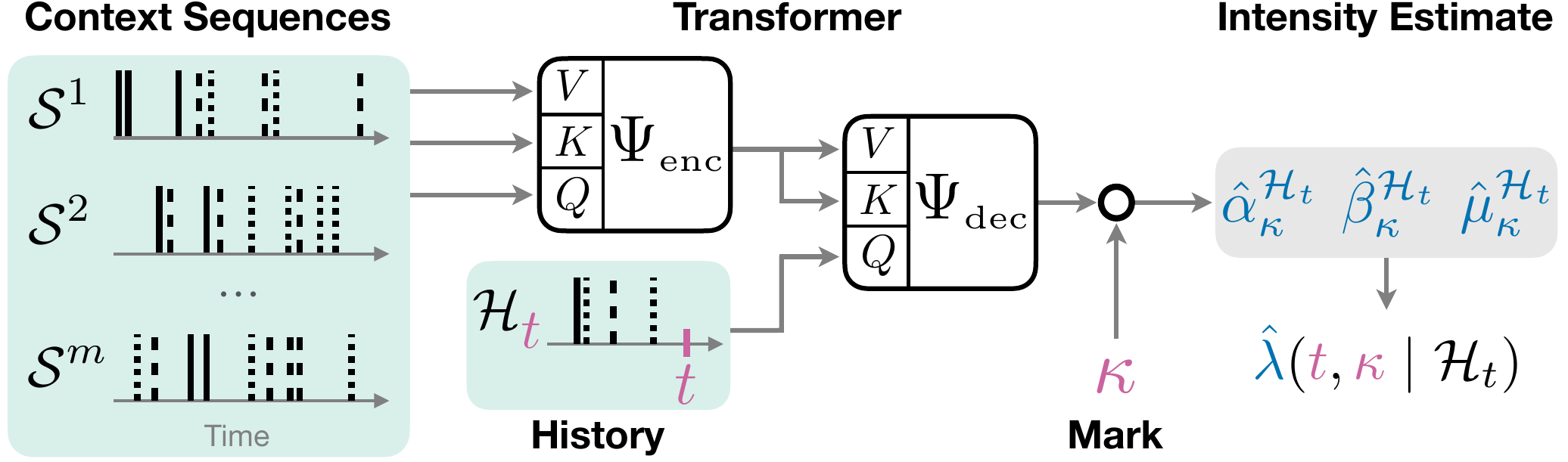}
\end{center}
\caption{
Schematic of \FIM \citep{fim_pp}. A \textit{context} of event sequences is encoded by a transformer encoder. The result is used as memory by a transformer decoder, which takes a specific event \textit{history} $\history_t$ as a query. The final embedding, combined with a target \textit{mark} $\markel$, is projected to parameters that define the \textit{conditional intensity function} $\hat{\lambda}(t, \markel \mid \history_t)$.
}
\label{fig:fim-pp-architecture}
\end{figure}

\FIM is a pretrained deep neural network that processes a \textit{context} of event sequences $\contextset = \{\contextseq^j\}_{j=1}^m$ to estimate the conditional intensity function $\hat{\lambda}$ that describes the observed dynamics. As shown in Figure \ref{fig:fim-pp-architecture}, the model uses an architecture that mimics the original transformer \citep{vaswani2017attention}.

First, each sequence in the context set $\contextset$ is embedded and processed by a transformer encoder to produce a summary vector. These summary vectors are then passed through another transformer encoder to share information across sequences, yielding a final context representation. To estimate the intensity at time $t$, an event history $\history_t$ is embedded and used as the query for a transformer decoder, which attends to the context representation. The output of the decoder is a context-aware history embedding $\histencfinal_t$.

This embedding is then used to parameterize a flexible intensity function for each mark $\markel^\prime$:
\begin{equation}
    \hat{\lambda}(t, \markel^\prime \mid \history_t) = \muhist + (\alphahist - \muhist) \exp\left(- \betahist (t - t_{n_\hist}) \right),
    \label{eq:intensity-parametrization}
\end{equation}
where $t_{n_\hist}$ is the time of the last event in the history, and the non-negative parameters $(\muhist, \alphahist, \betahist)$ are predicted by small MLPs from $\histencfinal_t$. The model is trained by minimizing the negative log-likelihood (NLL) of a target sequence, given the remaining sequences as context.

\section{Demonstration of Capabilities: Preliminary Results}
To demonstrate the effectiveness of our foundation model, we visualize its zero-shot performance on several synthetic temporal point processes. The following examples showcase the capabilities of the pretrained model, \FIM, applied directly to new data without any dataset-specific training. 

\begin{figure}[t]
\begin{center}
\includegraphics[width=\textwidth]{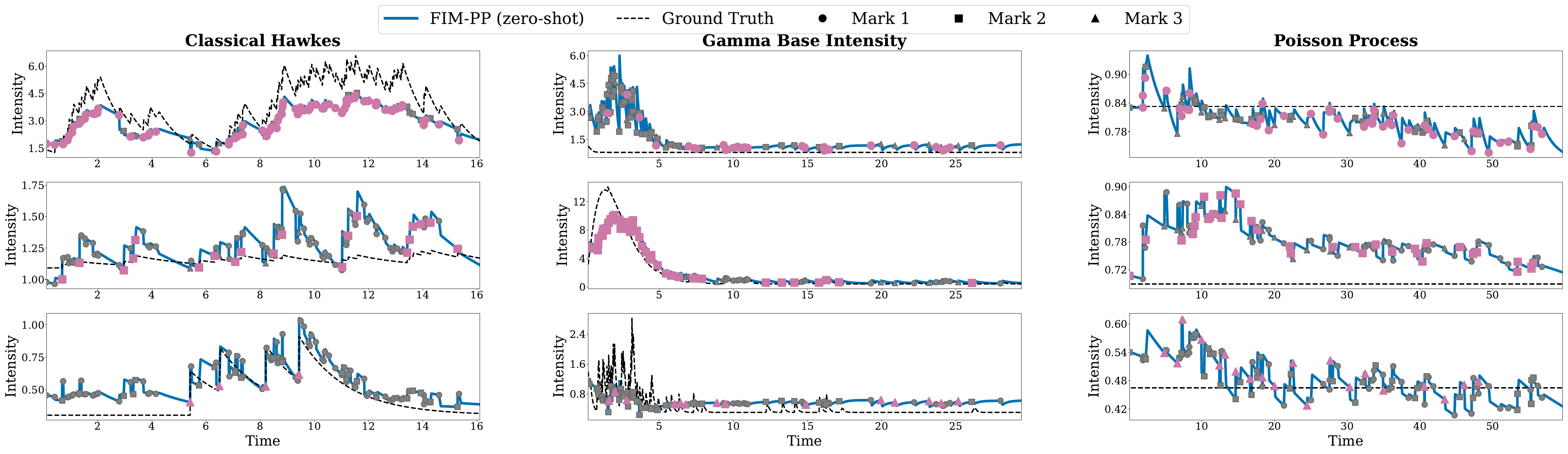}
\end{center}
\caption{
Demonstration of \FIM's \textit{zero-shot} inference capabilities on examples from three synthetic datasets: a synthetic Poisson process, a process with a constant base intensity, and one with a gamma-distributed base intensity. The zero-shot model effectively captures the underlying dynamics by leveraging its pretrained prior. The ground truth intensity is shown alongside the model's estimate, illustrating a close match even without dataset-specific training.
}
\label{fig:plots}
\end{figure}

Figure \ref{fig:plots} visually confirms the model's core ability to infer the conditional intensity of different underlying processes \textbf{in a zero-shot manner}. 
The plots show that \FIM produces remarkably accurate estimates that closely track the ground-truth dynamics for a variety of processes. This powerful generalization capability stems directly from its pretraining on a diverse corpus of synthetic Hawkes processes.

After downloading the model and providing it a few example sequences as context, a user can instantly obtain this intensity function for any given event history. This function can then be used for various downstream tasks, such as forecasting the time and type of the next event, simulating future event trajectories, or analyzing the excitatory and inhibitory influences between different event types. 
Evaluations of \FIM on real-world datasets can be found in \citep{fim_pp}.

\section{Conclusion}
\label{sec:conclusions}
In this work, we introduced \FIM, the first Foundation Inference Model capable of inferring MTPPs from real-world data. Our experiments demonstrated that a single \FIM, pretrained on synthetic Hawkes-process data only, can match the predictive performance of specialized MTPP methods in zero-shot mode. Finetuning on target data further boosts performance, enabling \FIM\ to outperform competing methods across the majority of evaluated tasks \cite{fim_pp}.

\textbf{Collaboration:} We are very interested in collaboration. We would be especially interested in solving real world problems with \FIM and encourage practitioners from any domain to reach out to us.

\section*{Acknowledgments}

This research has been funded by the Federal Ministry of Education and Research of Germany and the state of North-Rhine Westphalia as part of the Lamarr Institute for Machine Learning and Artificial Intelligence.

\bibliography{ais}

\end{document}